\documentclass[runningheads,a4paper]{llncs}
\usepackage{amsmath}
\usepackage{amssymb}
\usepackage{amsfonts}
\usepackage{amssymb}
\usepackage{graphicx}
\usepackage{breakcites}
\usepackage{color}
\usepackage{bm}
\usepackage{multirow}
\usepackage{booktabs}
\usepackage{soul}
\usepackage[raggedright]{sidecap}
\sidecaptionvpos{figure}{c} 
\usepackage{multirow}
\usepackage{floatrow}
\newfloatcommand{capbtabbox}{table}[][\FBwidth]
\usepackage{cite}
\definecolor{darkgreen}{rgb}{0,0.6,0.2}

\usepackage{url}
\newcommand{\keywords}[1]{\par\addvspace\baselineskip
\noindent\keywordname\enspace\ignorespaces#1}

\usepackage{subfigure}

\begin{document}

\mainmatter  

\title{Deep Kernelized Autoencoders}

\titlerunning{Deep Kernelized Autoencoders}

%
%
\author{Michael Kampffmeyer\inst{1}\thanks{michael.c.kampffmeyer@uit.no} \and Sigurd L{\o}kse\inst{1} \and Filippo M. Bianchi\inst{1} \and Robert Jenssen\inst{1} \and Lorenzo Livi\inst{2}
}
\authorrunning{M. Kampffmeyer et al.}

\institute{Machine Learning Group, UiT--The Arctic University of Norway\thanks{http://site.uit.no/ml/} \and Department of Computer Science, University of Exeter, UK}

%
%

\toctitle{Deep kernelized autoencoders}
\tocauthor{}
\maketitle

\begin{abstract}
In this paper we introduce the deep kernelized autoencoder, a neural network model that allows an explicit approximation of (i) the mapping from an input space to an arbitrary, user-specified kernel space and (ii) the back-projection from such a kernel space to input space. 
The proposed method is based on traditional autoencoders and is trained through a new unsupervised loss function. During training, we optimize both the reconstruction accuracy of input samples and the alignment between a kernel matrix given as prior and the inner products of the hidden representations computed by the autoencoder. Kernel alignment provides control over the hidden representation learned by the autoencoder.
Experiments have been performed to evaluate both reconstruction and kernel alignment performance. Additionally, we applied our method to emulate kPCA on a denoising task obtaining promising results.
\end{abstract}

\section{Introduction}
\label{sec:intro}

Autoencoders (AEs) are a class of neural networks that gained increasing interest in recent years~\cite{vincent2010stacked,kingma2013auto, santana2016information}. 
AEs are used for unsupervised learning of \emph{effective} hidden representations of input data ~\cite{Hinton504,bengio2009learning}. 
These representations should capture the information contained in the input data, while providing meaningful features for tasks such as clustering and classification \cite{6472238}.
However, what an \emph{effective} representation consists of is highly dependent on the target task.

In standard AEs, representations are derived by training the network to reconstruct inputs through either a bottleneck layer, thereby forcing the network to learn how to compress input information, or through an over-complete representation. 
In the latter, regularization methods are employed to, e.g., enforce sparse representations, make representations robust to noise, or penalize sensitivity of the representation to small changes in the input~\cite{6472238}.
However, regularization provides limited control over the nature of the hidden representation.

In this paper, we hypothesize that an \emph{effective} hidden representation should capture the relations among inputs, which are encoded in form of a kernel matrix. 
Such a matrix is used as a prior to be reproduced by inner products of the hidden representations learned by the AE. 
Hence, in addition to minimizing the reconstruction loss, we also minimize the normalized Frobenius distance between the prior kernel matrix and the inner product matrix of the hidden representations. 
We note that this process resembles the kernel alignment procedure \cite{wang2015overview}.

The proposed model, called \textit{deep kernelized autoencoder}, is related to recent attempts to bridge the performance gap between kernel methods and neural networks~\cite{wilson2016deep, NIPS2009_3628}. Specifically, it is connected to works on interpreting neural networks from a kernel perspective \cite{Montavon2011} and the Information Theoretic-Learning Auto-Encoder~\cite{santana2016information}, which imposes a prior distribution over the hidden representation in a variational autoencoder~\cite{kingma2013auto}.

In addition to providing control over the hidden representation, our method also has several benefits that compensate for important drawbacks of traditional kernel methods.
During training, we learn an explicit approximate mapping function from the input to a kernel space, as well as the associated back-mapping to the input space, through an end-to-end learning procedure. 
Once the mapping is learned, it can be used to relate operations performed in the approximated kernel space, for example linear methods (as is the case of kernel methods), to the input space. In the case of linear methods, this is equivalent to performing non-linear operations on the non-transformed data.
Mini-batch training is used in our proposed method in order to lower the computational complexity inherent to traditional kernel methods and, especially, spectral methods~\cite{scholkopf1998nonlinear,boser1992training,jenssen2010kernel}.
Additionally, our method applies to arbitrary kernel functions, even the ones computed through ensemble methods. To stress this fact, we consider in our experiments the probabilistic cluster kernel, a kernel function that is robust with regards to hyperparameter choices and has been shown to often outperform counterparts such as the RBF kernel~\cite{izquierdo2015spectral}.


\section{Background}
\label{sec:background}

\subsection{Autoencoders and stacked autoencoders}
\label{sec:autoencoders}

AEs simultaneously learn two functions.
The first one, \textit{encoder}, provides a mapping from an input domain, $\mathcal{X}$, to a code domain, $\mathcal{C}$, i.e., the hidden representation.
The second function, \textit{decoder}, maps from $\mathcal{C}$ back to $\mathcal{X}$.
For a single hidden layer AE, the encoding function $E(\cdot; \mathbf{W}_{E})$ and the decoding function $D(\cdot; \mathbf{W}_{D})$ are defined as
\begin{align}
    \begin{split}
    \label{eq:encoding_decoding}
    \mathbf{h} &= E(\mathbf{x}; \mathbf{W}_{E}) = \sigma(\mathbf{W}_E\mathbf{x} + \mathbf{b}_E) \\
    \mathbf{\tilde{x}} &= D(\mathbf{h}; \mathbf{W}_{D}) = \sigma(\mathbf{W}_D\mathbf{h} + \mathbf{b}_D),
    \end{split}
\end{align}
where $\sigma(\cdot)$ denotes a suitable transfer function (e.g., a sigmoid applied component-wise), $\mathbf{x}$, $\mathbf{h}$, and $\mathbf{\tilde{x}}$ denote, respectively, a sample from the input space, its hidden representation, and its reconstruction; finally, $\mathbf{W}_{E}$ and $\mathbf{W}_{D}$ are the weights and $\mathbf{b}_{E}$ and $\mathbf{b}_{D}$ the bias of the encoder and decoder, respectively.
For the sake of readability, we implicitly incorporate $\mathbf{b}_{E}, \mathbf{b}_{D}$ in the notation.
Accordingly, we can rewrite
\begin{equation}
    \label{eq:autoencoder}
    \mathbf{\tilde{x}} = D(E(\mathbf{x}; \mathbf{W}_{E}); \mathbf{W}_{D}).
\end{equation}

In order to minimize the discrepancy between the original data and its reconstruction, the parameters in Eq. \ref{eq:encoding_decoding} are typically learned by minimizing, usually through stochastic gradient descent (SGD), a reconstruction loss
\begin{equation}
    \label{eq:distortion}
    L_r(\mathbf{x}, \mathbf{\tilde{x}}) = \lVert \mathbf{x} - \mathbf{\tilde{x}} \rVert_{2}^{2} \; .
\end{equation}
%

Differently from Eq.~\ref{eq:encoding_decoding}, a stacked autoencoder (sAE) consists of several hidden layers \cite{Hinton504}. 
Deep architectures are capable of learning complex representations by transforming input data through multiple layers of nonlinear processing \cite{6472238}.
The optimization of the weights is harder in this case and pretraining is beneficial, as it is often easier to learn intermediate representations, instead of training the whole architecture end-to-end~\cite{bengio2009learning}. A very important application of pretrained sAE is the initialization of layers in deep neural networks~\cite{vincent2010stacked}.
Pretraining is performed in different phases, each of which consists of training a single AE. After the first AE has been trained, its encoding function $E(\cdot; \mathbf{W}_{E}^{(1)})$ is applied to the input and the resulting representation is used to train the next AE in the stacked architecture. Each layer, being trained independently, aims at capturing more abstract features by trying to reconstruct the representation in the previous layer. Once all individual AEs are trained, they are unfolded yielding a pretrained sAE. For a two-layer sAE, the encoding function consists of $E(E(\mathbf{x}; \mathbf{W}_{E}^{(1)});\mathbf{W}_{E}^{(2)})$, while the decoder reads $D(D(\mathbf{h}; \mathbf{W}_{D}^{(2)});\mathbf{W}_{D}^{(1)})$.
The final sAE architecture can then be fine-tuned end-to-end by back-propagating the gradient of the reconstruction error.

\subsection{A brief introduction to relevant kernel methods}
\label{sec:kernel_methods}

Kernel methods process data in a kernel space $\mathcal{K}$ associated with an input space $\mathcal{X}$ through an implicit (non-linear) mapping $\phi: \mathcal{X} \rightarrow \mathcal{K}$.
There, data are more likely to become separable by linear methods~\cite{cover1991elements}, which produces results that are otherwise only obtainable by nonlinear operations in the input space.
Explicit computation of the mapping $\phi(\cdot)$ and its inverse $\phi^{-1}(\cdot)$ is, in practice, not required. In fact, operations in the kernel space are expressed through inner products (kernel trick), which are computed as Mercer kernel functions in input space: $\kappa(\mathbf x_i, \mathbf x_j) = \langle \phi(\mathbf x_i), \phi(\mathbf x_j) \rangle$.

As a major drawback, kernel methods scale poorly with the number of data points $n$: traditionally, memory requirements of these methods scale with $\mathcal{O}(n^2)$ and computation with $\mathcal{O}(n^2\times d)$, where $d$ is the dimension~\cite{dai2014scalable}.
For example, kernel principal component analysis (kPCA)~\cite{scholkopf1998nonlinear}, a common dimensionality reduction technique that projects data into the subspace that preserves the maximal amount of variance in kernel space, requires to compute the eigendecomposition of a kernel matrix $\mathbf{K} \in \mathbb{R}^{n\times n}$, with $K_{ij}=\kappa(x_i, x_j), x_i, x_j\in\mathcal{X}$, yielding a computational complextiy $\mathcal{O}(n^3)$ and memory requirements that scale as $\mathcal{O}(n^2)$. For this reason, kPCA is not applicable to large-scale problems.
The availability of efficient (approximate) mapping functions, however, would reduce the complexity, thereby enabling these methods to be applicable on larger datasets~\cite{NIPS2009_3628}. Furthermore, by providing an approximation for $\phi^{-1}(\cdot)$, it would be possible to directly control and visualize data represented in $\mathcal{K}$.
Finding an explicit inverse mapping from $\mathcal{K}$ is a central problem in several applications, such as image denoising performed with kPCA, also known as the pre-image problem~\cite{bakir2004learning,honeine2011closed}.

\subsection{Probabilistic Cluster Kernel}
\label{sec:pck}

The Probabilistic Cluster Kernel (PCK) \cite{izquierdo2015spectral} adapts to inherent structures in the data and it does not depend on any critical user-specified hyperparameters, like the width in Gaussian kernels. The PCK is trained by fitting multiple Gaussian Mixture Models (GMMs) to input data and then combining these models into a single kernel.
In particular, GMMs are trained for a variety of mixture components $g = 2, 3, \ldots, G$, each with different randomized initial conditions $q = 1, 2, \ldots, Q$.
Let $\boldsymbol \pi_i(q, g)$ denote the \textit{posterior distribution} for data point $\mathbf x_i$ under a GMM with $g$ mixture components and initial condition $q$. The PCK is then defined as
\begin{equation}\label{eq:pck}
    \kappa_{\mathrm{PCK}}(\mathbf x_i, \mathbf x_j) 
    = \frac{1}{Z} \sum_{q = 1}^Q \sum_{g = 2}^G
    \boldsymbol \pi_i^T(q, g) \boldsymbol \pi_j (q, g),
\end{equation}
where $Z$ is a normalizing constant.

Intuitively, the posterior distribution under a mixture model contains probabilities that a given data point belongs to a certain mixture component in the model.
Thus, the inner products in Eq. \ref{eq:pck} are large if data pairs often belong to the same mixture component.
By averaging these inner products over a range of $G$ values, the kernel function has a large value only if these data points are similar on both global scale (small $G$) and local scale (large $G$).

\section{Deep kernelized autoencoders}
\label{sec:kernelizedae}

\begin{figure}[t!]
  \centering
\includegraphics[width=0.6\textwidth, keepaspectratio]{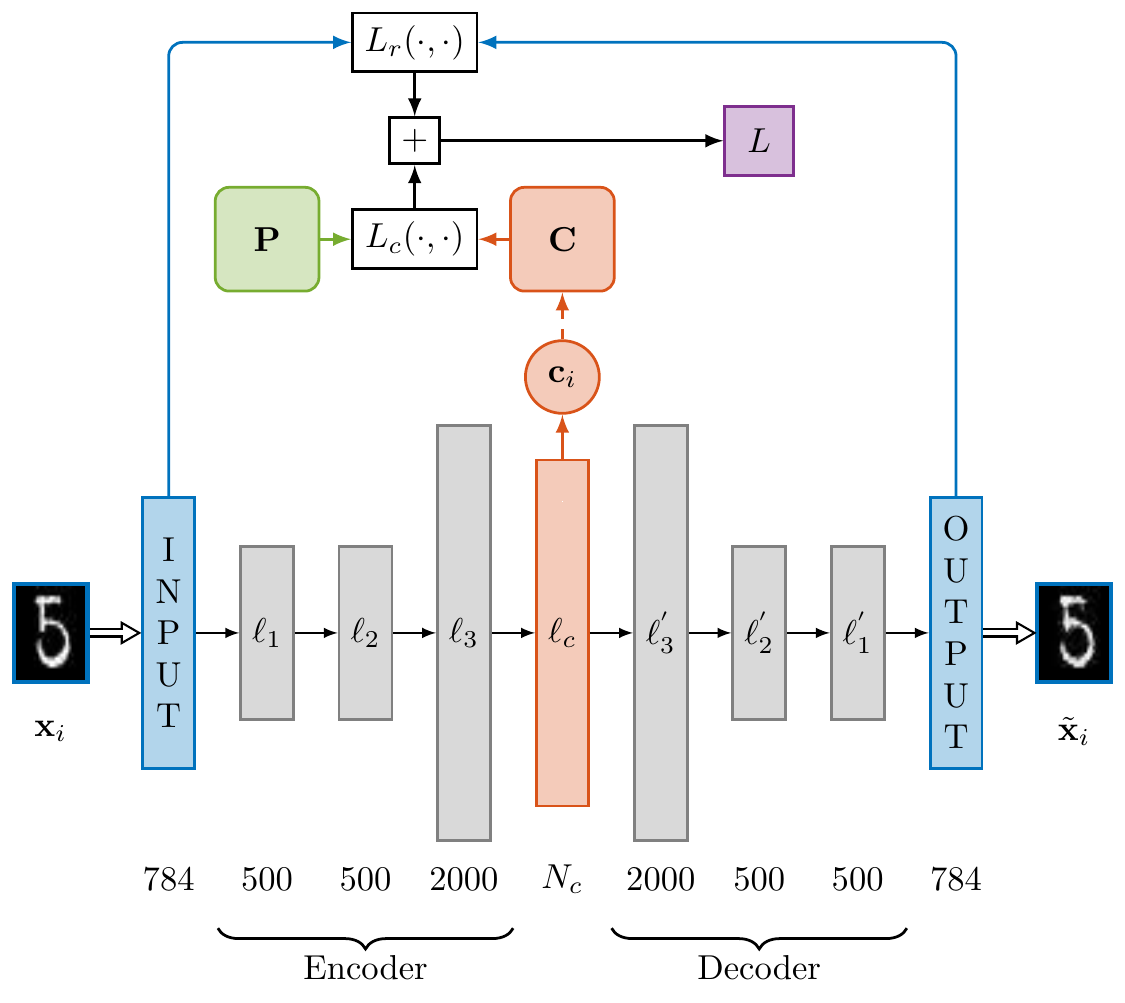}
  \caption{Schematic illustration of dkAE architecture. Loss function $L$ depends on two terms. First, $L_r(\cdot,\cdot)$, is the reconstruction error between true input $\mathbf{x}_i$ and output of dkAE, $\tilde{\mathbf{x}}_i$. Second term, $L_c(\cdot, \cdot)$, is the distance measure between matrices $\mathbf{C}$ (computed as inner products of codes $\{ \mathbf{c}_i \}_{i=1}^{n}$) and the target prior kernel matrix $\mathbf{P}$. For mini-batch training matrix $\mathbf{C}$ is computed over the codes of the data in the mini-batch and that distance is compared to the submatrix of $\mathbf{P}$ related to the current mini-batch.}
	\label{fig:kAE_arch}
\end{figure}

In this section, we describe our contribution, which is a method combining AEs with kernel methods: the deep kernelized AE (dkAE).
A dkAE is trained by minimizing the following loss function
\begin{equation}
    \label{eq:cost}
    L = (1-\lambda) L_r(\mathbf{x}, \mathbf{\tilde{x}}) + \lambda L_c(\mathbf{C}, \mathbf{P}),
\end{equation}
where $L_r(\cdot, \cdot)$ is the reconstruction loss in Eq.~\ref{eq:distortion}.
$\lambda$ is a hyperparameter ranging in $[0, 1]$, which weights the importance of the two objectives in Eq.~\ref{eq:cost}. 
For $\lambda=0$, the loss function simplifies to the traditional AE loss in Eq.~\ref{eq:autoencoder}.
$L_c(\cdot, \cdot)$ is the code loss, a distance measure between two matrices, $\mathbf{P} \in \mathbb{R}^{n \times n}$, the kernel matrix given as prior, and $\mathbf{C} \in \mathbb{R}^{n \times n}$, the 
inner product matrix of codes associated to input data.
The objective of $L_c(\cdot,\cdot)$ is to enforce the similarity between $\mathbf{C}$ and the prior $\mathbf{P}$. 
A depiction of the training procedure is reported in Fig. \ref{fig:kAE_arch}.

We implement $L_c(\cdot, \cdot)$ as the normalized Frobenius distance between $\mathbf{C}$ and $\mathbf{P}$. Each matrix element $C_{ij}$ in $\mathbf{C}$ is given by 
$C_{ij}=E(\mathbf{x}_i) \cdot E(\mathbf{x}_j)$ and the code loss is computed as
\begin{equation}
\label{eq:regularization}
    L_c(\mathbf{C}, \mathbf{P}) = \Bigg{\lVert} \frac{\mathbf{C}}{\|\mathbf{C}\|_F} - \frac{\mathbf{P}}{\|\mathbf{P}\|_F} \Bigg{\rVert}_{F}.
\end{equation}
    
Minimizing the normalized Frobenius distance between the kernel matrices is equivalent to maximizing the traditional kernel alignment cost, since
\begin{equation} \label{eq:normdistance}
   \Bigg{\lVert} \frac{\mathbf{C}}{\|\mathbf{C}\|_F} - \frac{\mathbf{P}}{\|\mathbf{P}\|_F} \Bigg{\rVert}_{F} =
   \sqrt{2 - 2A(\mathbf C, \mathbf P)},
\end{equation} 
where $A(\mathbf C, \mathbf P) = \frac{\langle \mathbf C, \mathbf P \rangle_F}
{\|\mathbf C\|_F \|\mathbf P\|_F}$ is exactly the kernel alignment cost function \cite{christianini2001kernel,wang2015overview}.
Note that the distance in Eq. \ref{eq:normdistance} can be implemented also with more advanced differentiable measures of (dis)similarity between PSD matrices, such as divergence and mutual information \cite{kulis2009low,giraldo2015measures}. However, these options are not explored in this paper and are left for future research.

In this paper, the prior kernel matrix $\mathbf{P}$ is computed by means of the PCK algorithm introduced in Section \ref{sec:pck}, such that $\mathbf{P}=\mathbf{K}_{\text{PCK}}$. 
However, our approach is general and \textit{any} kernel matrix can be used as prior in Eq. \ref{eq:regularization}.

\subsection{Mini-batch training}
We use mini batches of $k$ samples to train the dkAE, thereby avoiding the computational restrictions of kernel and especially spectral methods outlined in Sec.~\ref{sec:kernel_methods}. 
Making use of mini-batch training, the memory complexity of the algorithm can be reduced to $O(k^2)$, where $k \ll n$. Finally, we note that the computational complexity scales linearly with regards to the parameters in the network.
In particular, given a mini batch of $k$ samples, the dkAE loss function is defined by taking the average of the per-sample reconstruction cost
\begin{equation}
\label{eq:cost_minibatch}
    L_{\mathrm{batch}}=\frac{1-\lambda}{kd} \sum_{i=1}^{k} L_r(\mathbf{x}_i, \mathbf{\tilde{x}}_i)  + \lambda
   \Bigg{\lVert} \frac{\mathbf{C}_k}{\|\mathbf{C}_k\|_F} - \frac{\mathbf{P}_k}{\|\mathbf{P}_k\|_F} \Bigg{\rVert}_{F},
\end{equation}
where $d$ is the dimensionality of the input space, $\mathbf{P}_k$ is a subset of $\mathbf{P}$ that contains only the $k$ rows and columns related to the current mini-batch, and $\mathbf{C}_k$ contains the inner products of the codes for the specific mini-batch. Note that $\mathbf{C}_k$ is re-computed in each mini batch.

\subsection{Operations in code space}
\label{sec:lin_op}
\begin{figure}[tbp]
  \centering
  \includegraphics[width=0.62\textwidth, keepaspectratio]{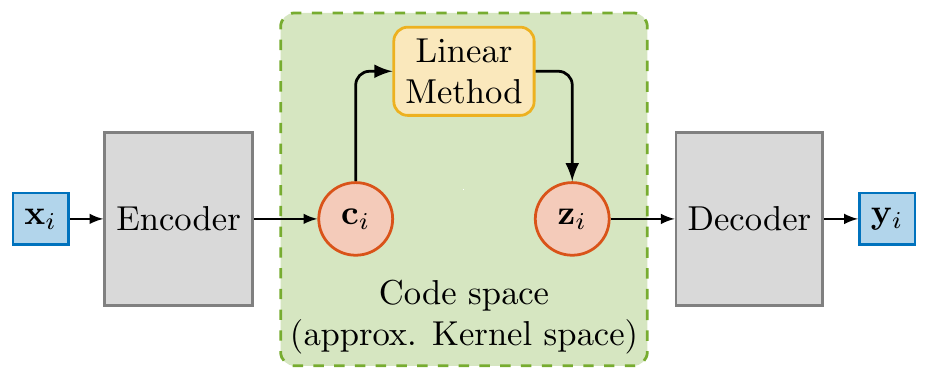}
  \caption{The encoder maps input $\mathbf{x}_i$ to $\mathbf{c}_i$, which lies in code space. In dkAEs, the code domain approximates the space associated to the prior kernel $\mathbf{P}$. A linear method receives input $\mathbf{c}_i$ and produces output $\mathbf{z}_i$.
  The decoder maps $\mathbf{z}_i$ back to input space. The result $\mathbf{y}_i$ can be seen as the output of a non-linear operation on $\mathbf{x}_i$ in input space.}
	\label{fig:kAE_method}
\end{figure}

Linear operations in code space can be performed as shown in Fig. \ref{fig:kAE_method}.
The encoding scheme of the proposed dkAE explicitly approximates the function $\phi(\cdot)$ that maps an input $\mathbf{x}_i$ onto the kernel space.
In particular, in a dkAE the feature vector $\phi(\mathbf{x}_i)$ is approximated by the code $\mathbf{c}_i$.
Following the underlying idea of kernel methods and inspired by Cover's theorem~\cite{cover1991elements}, which states that a high dimensional embedding is more likely to be linearly separable, linear operations can be performed on the code.
A linear operation on $\mathbf{c}_i$ produces a result in the code space, $\mathbf{z}_i$, relative to the input $\mathbf{x}_i$.
Codes are mapped back to the input space by means of a decoder, which in our case approximates the inverse mapping $\phi(\cdot)^{-1}$ from the kernel space back to the input domain. Unlike other kernel methods where this explicit mapping is not defined, this fact permits visualization and interpretation of the results in the original space.

\section{Experiments and results}
\label{sec:experiments}

In this section, we evaluate the effectiveness of dkAEs on different benchmarks.
In the first experiment we evaluate the effect on the two terms of the objective function (Eq. \ref{eq:cost_minibatch}) when varying the hyperparameters $\lambda$ (in Eq. \ref{eq:cost}) and the size of the code layer.
In a second experiment, we study the reconstruction and the kernel alignment. 
Further we compare dkAEs approximation accuracy of the prior kernel matrix to kPCA as the number of principle components increases. 
Finally, we present an application of our method for image denoising, where we apply PCA in the dkAE code space $\mathcal{C}$ to remove noise.

For these experiments, we consider the MNIST dataset, consisting of $60000$ images of handwritten digits. However, we use a subset of $20000$ samples due to the computational restrictions imposed by the PCK, which we use to illustrate dkAEs ability to learn arbitrary kernels, even if they originate from an ensemble procedure.
We train the PCK by fitting the GMMs on a subset of 200 training samples, with the parameters $Q=G=30$.
Once trained, the GMM models are applied on the remaining data to calculate the kernel matrix.
We use 70\%,15\% and 15\% of the data for training, validation, and testing, respectively.

\subsection{Implementation}
The network architecture used in the experiments is $d-500-500-2000-N_c$ (see Fig. \ref{fig:kAE_arch}), which has been demonstrated to perform well on several datasets, including MNIST, for both supervised and unsupervised tasks~\cite{maaten2009learning,hinton2006fast}.
Here, $N_c$ refers to the dimensionality of the code layer. Training was performed using the sAE pretraining approach outlined in Sec.~\ref{sec:autoencoders}.
To avoid learning the identify mapping on each individual layer, we applied a common \cite{kamyshanska2015potential} regularization technique where the encoder and decoder weights are tied, i.e., $W_{E} = W_{D}^T$. This is done during pretraining and fine-tuning. 
Unlike in traditional sAEs, to account for the kernel alignment objective, the code layer is optimized according to Eq.~\ref{eq:cost} \textit{also} during pretraining.

Size of mini-batches for training was chosen to be $k=200$ randomly, independently sampled data points; in our experiments, an epoch consists of processing $(n/k)^2$ batches.
Pretraining is performed for $30$ epochs per layer and the final architecture is fine-tuned for $100$ epochs using gradient descent based on Adam~\cite{kingma2014adam}. 
The dkAE weights are randomly initialized according to Glorot et al.~\cite{Glorot10understandingthe}.

\subsection{Influence of hyperparameter $\lambda$ and size $N_c$ of code layer}
\label{sec:exp1}
\begin{figure}[tbp]
\centering
\subfigure[]{
\includegraphics[width=0.44\textwidth]{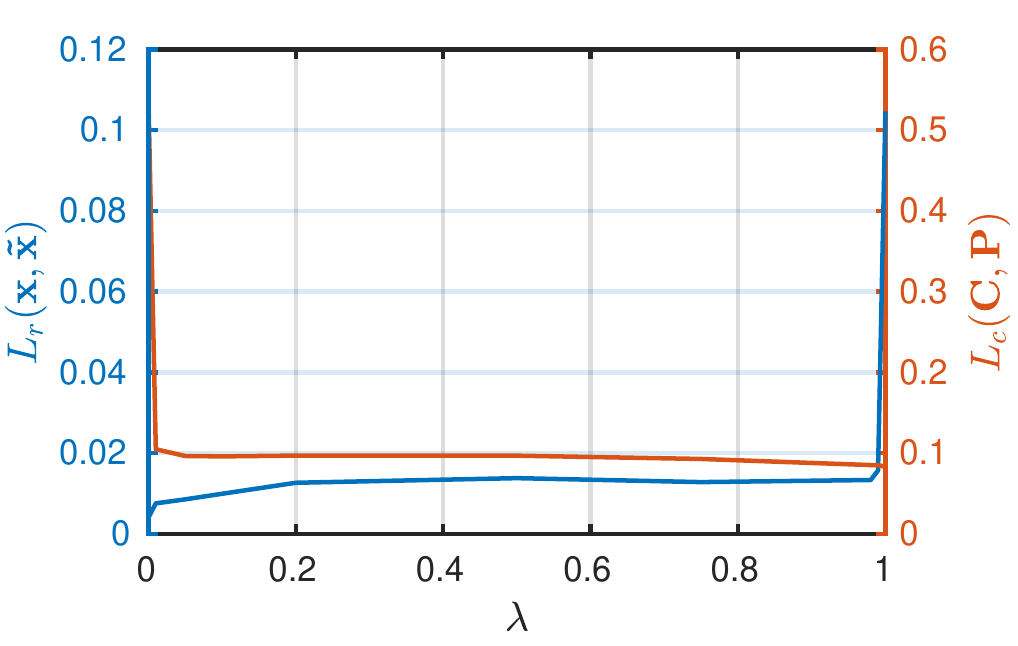}
\label{fig:lambda_experiment}}\hspace{-1em}
\subfigure[]{
\includegraphics[width=0.44\textwidth]{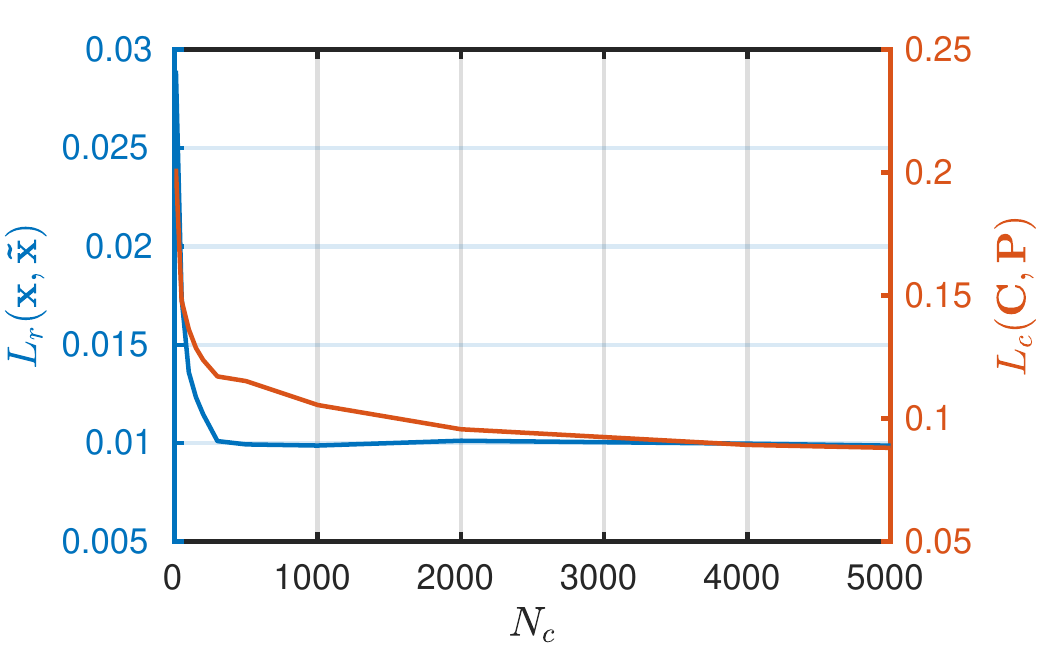}
\label{fig:nc_experiment}}\hspace{-1em}
\caption{(a): Tradeoff when choosing $\lambda$. High $\lambda$ values result in low $L_c$, but high reconstruction cost, and vice-versa. (b): Both $L_c$ and reconstruction costs decrease when code dimensionality $N_c$ increases.}
\label{fig:mnist_map}
\end{figure}

In this experiment, we evaluate the influence of the two main hyperparameters that determine the behaviour of our architecture.
Note that the experiments shown in this section are performed by training the dkAE on the training set and evaluating the performance on the validation set. We evaluate both the out-of-sample reconstruction $L_r$ and $L_c$.
Figure~\ref{fig:lambda_experiment} illustrates the effect of $\lambda$ for a fixed value $N_c=2000$ of neurons in the code layer. It can be observed that the reconstruction loss $L_r$ increases as more and more focus is put on minimizing $L_c$ (obtained by increasing $\lambda$). This quantifies empirically the trade-off in optimizing the reconstruction performance and the kernel alignment at the same time.
Similarly, it can be observed that $L_c$ decreases when increasing $\lambda$. By inspecting the results, specifically the near constant losses for $\lambda$ in range [0.1,0.9] the method appears robust to changes in hyperparameter $\lambda$.

Analyzing the effect of varying $N_c$ given a fixed $\lambda=0.1$ (Figure~\ref{fig:nc_experiment}), we observe that both losses decrease as $N_c$ increases. 
This could suggest that an even larger architecture, characterized by more layers and more neurons w.r.t. the architecture adopted, might work well, as the dkAE does not seem to overfit, due also to the regularization effect provided by the kernel alignment.

\subsection{Reconstruction and kernel alignment}
\label{sec:exp1b}
\begin{figure}[tbp]
\centering
\minipage[t]{0.22\textwidth}
  \includegraphics[width=\linewidth]{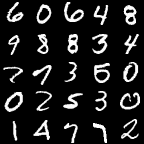}
  \centering{Original}
\endminipage\hspace{0.2cm}
\minipage[t]{0.22\textwidth}%
  \includegraphics[width=\linewidth]{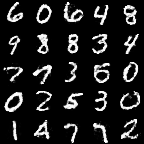}
  \centering{$\lambda=0.75$}
\endminipage\hspace{0.2cm}
\minipage[t]{0.22\textwidth}
  \includegraphics[width=\linewidth]{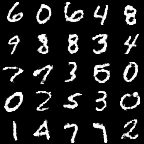}
  \centering{$\lambda=0.1$}
\endminipage\\
\minipage[t]{0.22\textwidth}
  \includegraphics[width=\linewidth]{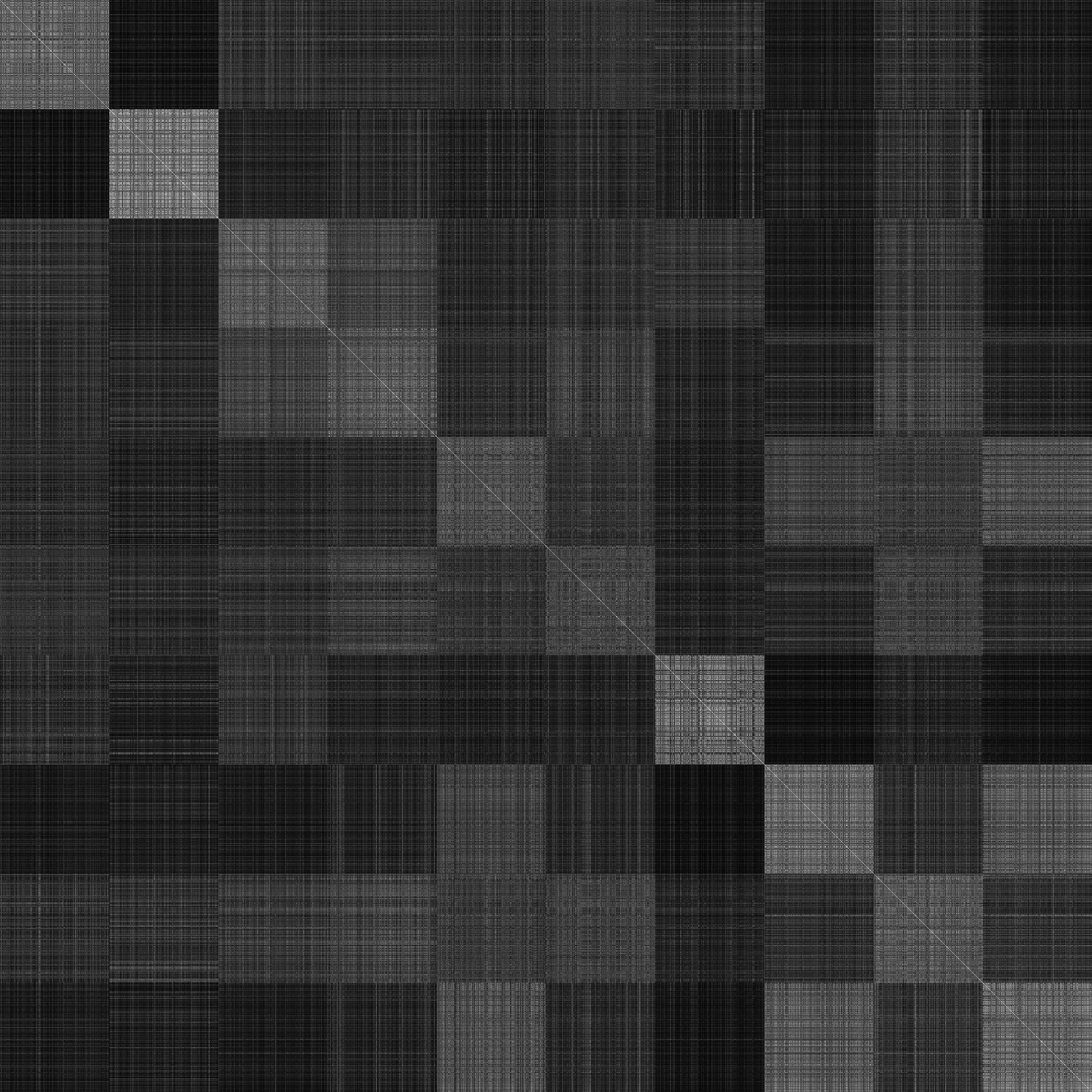}
  \centering{Prior}
\endminipage\hspace{0.2cm}
\minipage[t]{0.22\textwidth}
  \includegraphics[width=\linewidth]{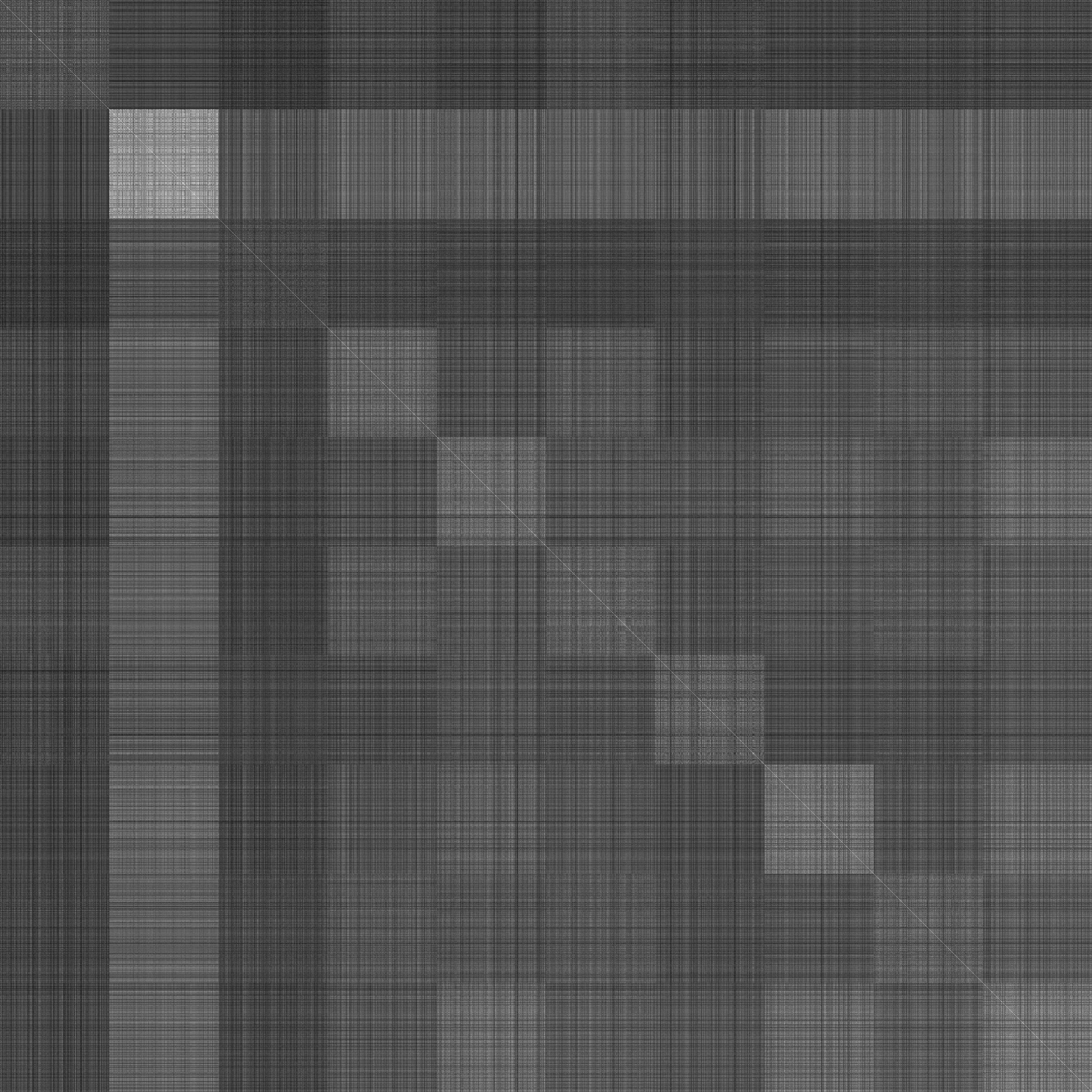}
  \centering{$\lambda=0.0$}
\endminipage\hspace{0.2cm}
\minipage[t]{0.22\textwidth}%
  \includegraphics[width=\linewidth]{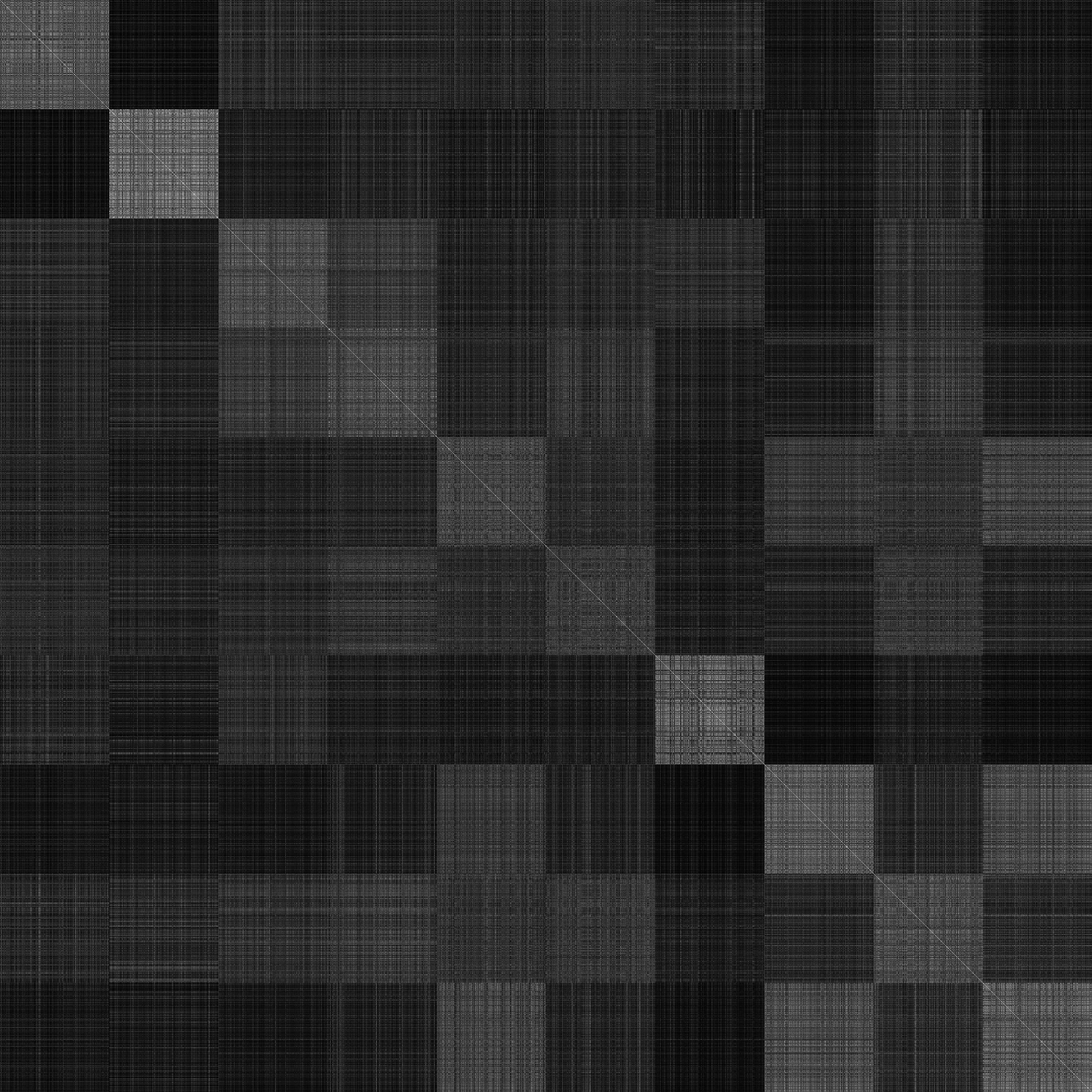}
  \centering{$\lambda=0.1$}
\endminipage
\caption{Illustrating the reconstruction and kernel alignment trade-off for different $\lambda$ values. We note that the reconstruction for a small $\lambda$ is generally better (see also Figure~\ref{fig:lambda_experiment}), but that small $\lambda$ yields high $L_c$.}\label{fig:lambda_experiment_reconstruction}
\end{figure}

According to the previous results, in the following experiments we set $\lambda=0.1$ and $N_c=2000$.
Figure~\ref{fig:lambda_experiment_reconstruction} illustrates the results in Sec.~\ref{sec:exp1} qualitatively by displaying a set of original images from our test set and their reconstruction for the chosen $\lambda$ value and a  non-optimal one. 
Similarly, the prior kernel (sorted by class in the figure, to ease the visualization) and the dkAEs approximated kernel matrices, relative to test data, are displayed for two different $\lambda$ values. 
Notice that, to illustrate the difference with a traditional sAE, one of the two $\lambda$ values is set to zero. 
It can be clearly seen that, for $\lambda=0.1$, both the reconstruction and the kernel matrix, resemble the original closely, which agrees with the plots in Figure~\ref{fig:lambda_experiment}.

Inspecting the kernels obtained in Figure~\ref{fig:lambda_experiment_reconstruction}, we compare the distance between the kernel matrices, $\mathbf{C}$ and $\mathbf{P}$, and the ideal kernel matrix, obtained by considering supervised information. We build the ideal kernel matrix $\mathbf{K}_I$, where $K_I(i,j) = 1$ if elements $i$ and $j$ belong to same class, otherwise $K_I(i,j) = 0$. 
Table~\ref{tab:results_ideal} illustrates that the kernel approximation produced by dkAE outperforms a traditional sAE with regards to kernel alignment with the ideal kernel. Additionally it can be seen that the kernel approximation actually improves slightly on the kernel prior, which we hypothesise is due to the regularization that is imposed by the reconstruction objective.

\bgroup
\def\arraystretch{1} 
\setlength\tabcolsep{1em} 
  \begin{table*}[tbp]\scriptsize\centering
  \caption{Computing $L_c$ with respect to an ideal kernel matrix $\mathbf{K}_I$ for our test dataset ($10$ classes) and comparing relative improvement for the three kernels in Figure~\ref{fig:lambda_experiment_reconstruction}. Prior kernel $\mathbf{P}$, a traditional sAE ($\lambda=0$) $\mathbf{K}_{AE}$, and dkAEs $\mathbf{C}$.}
  \vspace{-0.05cm}
  \begin{tabular}{c|c|ccc}
    \bottomrule[1.5pt]
    \multirow{2}{*}{\textbf{Kernel}} & \multirow{2}{*}{$L_c(\cdot, \mathbf{K}_I)$} & \multicolumn{3}{c}{Improvement [\%]~vs.} \\
    & & $\mathbf{P}$ & $\mathbf{K}_{AE}$ & $\mathbf{C}$ \\
    \toprule[1.5pt]
    $\mathbf{P}$ & 1.0132 & 0 & 12.7  & -0.2 \\ 
    $\mathbf{K}_{AE}$ & 1.1417 & -11.3 & 0 & -11.4 \\ 
    $\mathbf{C}$ & 1.0115 & 0.2 & 12.9 & 0 \\ 
    \toprule[1.5pt]
  \end{tabular}
  \label{tab:results_ideal}
  \end{table*}
\egroup

\subsection{Approximation of kernel matrix given as prior}
\label{sec:exp2}
In order to quantify the kernel alignment performance, we compare dkAE to the approximation provided by kPCA when varying the number of principal components. 
For this test, we take the kernel matrix $\mathbf{P}$ of the training set and compute its eigendecomposition. We then select an increasing number of components $m$ (with $m\geq1$ components associated with the largest eigenvalues) to project the input data: $\mathbf{Z}_m = \mathbf{E}_m \bm{\Lambda}_{m}^{1/2}, d=2, ..., N$. 
The approximation of the original kernel matrix (prior) is then given as $\mathbf{K}_m = \mathbf{Z}_m \mathbf{Z}_{m}^{T}$. We compute the distance between $\mathbf{K}_m$ and $\mathbf{P}$ following Eq. \ref{eq:normdistance} and compare it to dissimilarity between $\mathbf{P}$ and $\mathbf{C}$. 
For evaluating the out-of-sample performance, we use the Nystr{\"o}m approximation for kPCA~\cite{scholkopf1998nonlinear} and compare it to the dkAE kernel approximation on the test set.

\begin{SCfigure}[1.4][tbp]
  \centering
\includegraphics[width=0.6\textwidth, keepaspectratio, trim={0.7cm 0.1cm 0.7cm 0.1cm},clip]{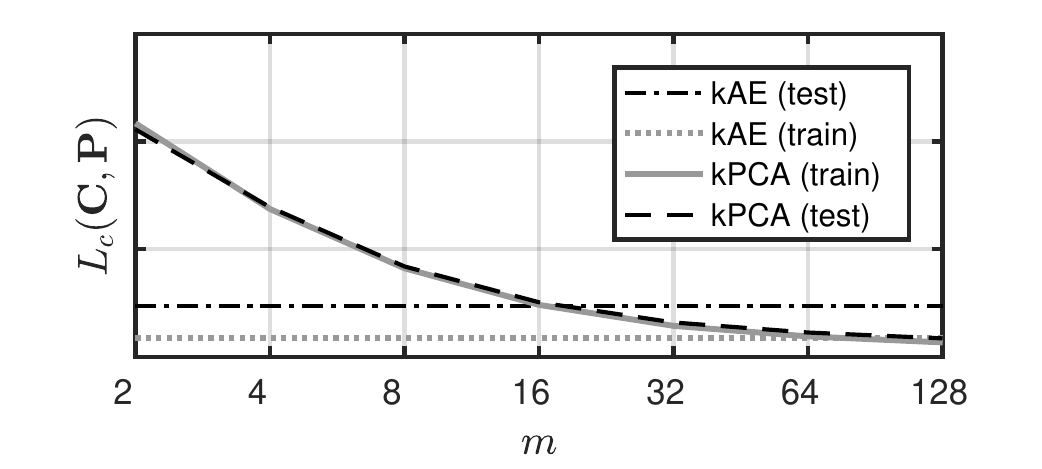}
  \caption{Comparing dkAEs approximation of the kernel matrix to kPCA for an increasing number of components. The plot shows that dkAE reconstruction is more accurate for low number (i.e., $m<16$) of components.}
	\label{fig:kPCA_exp}
\end{SCfigure}

Figure \ref{fig:kPCA_exp} shows that the approximation obtained by means of dkAEs outperforms kPCA when using a small number of components, i.e., $m<16$. Note that it is common in spectral methods to chose a number of components equal to the number of classes in the dataset~\cite{ng2001spectral} in which case, for the $10$ classes in the MNIST dataset, dkAE would outperform kPCA. As the number of selected components increases, the approximation provided by kPCA will perform better. However, as shown in the previous experiment (Sec. \ref{sec:exp1b}), this does not mean that the approximation performs better with regards to the ideal kernel. In fact, in that experiment the kernel approximation by dkAE actually performed at least as well as the prior kernel (kPCA with all components taken into account).

\subsection{Linear operations in code space}
Here we hint at the potential of performing operations in code space as described in Sec.~\ref{sec:lin_op}. We try to emulate kPCA by performing PCA in our learned kernel space and evaluate the performance on the task of denoising. Denoising is a task that requires both a mapping to the kernel space, as well as a back-projection. For traditional kernel methods no explicit back-projection exists, but approximate solutions to this so called pre-image problem have been proposed~\cite{bakir2004learning,honeine2011closed}. We chose the method proposed by Bakir et al.~\cite{bakir2004learning}, where they use kernel ridge regression, such that a different kernel (in our case an RBF) can be used for the back-mapping. As it was a challenging to find a good $\sigma$ for the RBF kernel that captures all numbers in the MNIST dataset, we performed this test on the $5$ and $6$ class only. The regularization parameter and the $\sigma$ required for the back-projection where found via grid search, where the best regularization parameter (according to MSE reconstruction) was found to be $0.5$ and $\sigma$ as the median of the euclidean distances between the projected feature vectors.

\begin{figure}[tbp]
\begin{floatrow}
\capbtabbox{%
     \bgroup
        \scriptsize\centering
        \def\arraystretch{1.4} 
        \setlength\tabcolsep{1em} 
          \begin{tabular}{ccc}
            \bottomrule[1.5pt]
            \textbf{Noise std.} & \textbf{kPCA} & \textbf{dkAE+PCA}\\
            \toprule[1.5pt]
             0.25 & 0.0427 & 0.0358  \\ 
            \toprule[1.5pt]
          \end{tabular}
        \egroup
}{%
  \caption{Mean squared error for reconstruction.}%
  \label{tab:denoising_res}
}
\ffigbox{%
    \includegraphics[width=0.3\linewidth]{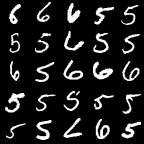}%
    \hspace{0.2cm}
    \includegraphics[width=0.3\linewidth]{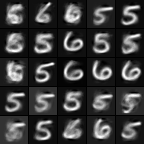}%
    \hspace{0.2cm}
    \includegraphics[width=0.3\linewidth]{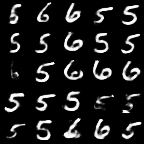}
}{%
  \caption{Original images (left), the reconstruction with kPCA (center) and with dkAE+PCA (right).}%
  \label{fig:denoising_res}
}
\end{floatrow}
\end{figure}

Both models are fitted on the training set and Gaussian noise is added to the test set. For both methods $32$ principle components are used. Tab.~\ref{tab:denoising_res} illustrates that dkAE+PCA outperforms kPCAs reconstruction with regards to mean squared error. However, as this is not necessarily a good measure for denoising~\cite{bakir2004learning}, we also visualize the results in Fig.~\ref{fig:denoising_res}. It can be seen that dkAE yields sharper images in the denoising task.

\section{Conclusions}
\label{sec:conclusions}

In this paper, we proposed a novel model for autoencoders, based on the definition of a particular unsupervised loss function. The proposed model enables us to learn an approximate embedding from an input space to an arbitrary kernel space as well as the projection from the kernel space back to input space through an end-to-end trained model. It is worth noting that, with our method, we are able to approximate arbitrary kernel functions by inner products in the code layer, which allows us to control the representation learned by the autoencoder. In addition, it enables us to emulate well-known kernel methods such as kPCA and scales well with the number of data points.

A more rigorous analysis of the learned kernel space embedding, as well as applications of the code space representation for clustering and/or classification tasks, are left as future works.

\subsubsection*{Acknowledgments.} We gratefully acknowledge the support of NVIDIA Corporation with the donation of the GPU used for this research. This work was partially funded by the Norwegian Research Council FRIPRO grant no.\ 239844 on developing the \emph{Next Generation Learning Machines}.

\bibliographystyle{splncs03}
\bibliography{bibliography.bib}
\clearpage

\end{document}